 \documentclass[runningheads]{llncs}
\usepackage{graphicx}
\usepackage{amsmath,amssymb} 
\usepackage{color}
\usepackage[width=122mm,left=12mm,paperwidth=146mm,height=193mm,top=12mm,paperheight=217mm]{geometry}

\usepackage{times}
\usepackage{epsfig}
\usepackage{graphicx}
\usepackage{amsmath}
\usepackage{amssymb}
\usepackage{multirow}
\usepackage{subcaption}
\usepackage{verbatim}
\usepackage{algorithm}
\usepackage{algorithmic}
\usepackage{hhline}

\usepackage[pagebackref=true,breaklinks=true,letterpaper=true,colorlinks,bookmarks=false]{hyperref}

\makeatletter
\usepackage{xspace}
\def\@onedot{\ifx\@let@token.\else.\null\fi\xspace}
\DeclareRobustCommand\onedot{\futurelet\@let@token\@onedot}

\newcommand{\cI}{{\mathcal{I}}}

\begin{document}
\pagestyle{headings}
\mainmatter
\def\ECCV16SubNumber{1170}  


\title{Exploiting Semantic Information and Deep Matching for Optical Flow}

\titlerunning{Exploiting Semantic Information and Deep Matching for Optical Flow}

\authorrunning{Min Bai$^\star$, Wenjie Luo$^\star$, Kaustav Kundu, Raquel Urtasun}

\author{Min Bai\thanks{Denotes equal contribution}, Wenjie Luo$^\star$, Kaustav Kundu, Raquel Urtasun\\
{\normalsize Department of Computer Science, University of Toronto}\\
{\tt \small \{mbai, wenjie, kkundu, urtasun\}@cs.toronto.edu}}
\institute{}

\maketitle

\begin{abstract}
We tackle the problem of estimating optical flow from a monocular camera in the context of autonomous driving. 
We build on the observation that the scene is typically composed of a static background, as well as a relatively small number of traffic participants which move rigidly in 3D. We propose to estimate the traffic participants using instance-level segmentation. For each traffic participant, we use the epipolar constraints that govern each independent motion for faster and more accurate estimation. 
Our second contribution is a new convolutional net that learns to perform flow matching, and  is able to estimate the uncertainty of its matches. This is a core element of our flow estimation pipeline.
We demonstrate the effectiveness of our approach in the challenging KITTI 2015 flow benchmark, and show that our approach outperforms published approaches by a large margin.

\end{abstract}


\section{Introduction}

Despite many decades of research, estimating dense optical flow is still an open problem. Large displacements, texture-less regions, specularities, shadows, and strong changes in illumination continue to pose difficulties. Furthermore,  flow estimation is computationally very demanding,  as the typical range for a pixel's potential motion can contain more than 30K possibilities. This poses   many problems for discrete methods, therefore most recent methods rely on continuous optimization \cite{Revaud,Black2013}. 

In this paper, we are interested in computing optical flow in the context of autonomous driving. We argue that strong priors can be exploited in this context to make estimation more robust (and potentially faster). In particular, we build on the observation that the scene is typically composed of a static background, as well as a relatively small number of traffic participants which move rigidly in 3D.
To exploit such intuition, we need to reliably identify the independently moving objects, and estimate their motion. Past methods typically attempt to segment the objects based solely on motion. However, this is a chicken and egg problem: an accurate motion estimation is necessary for an accurate motion segmentation, yet the latter also circularly depends upon the former. 

\begin{figure*}[t]
\begin{center}
\includegraphics[width=\linewidth]{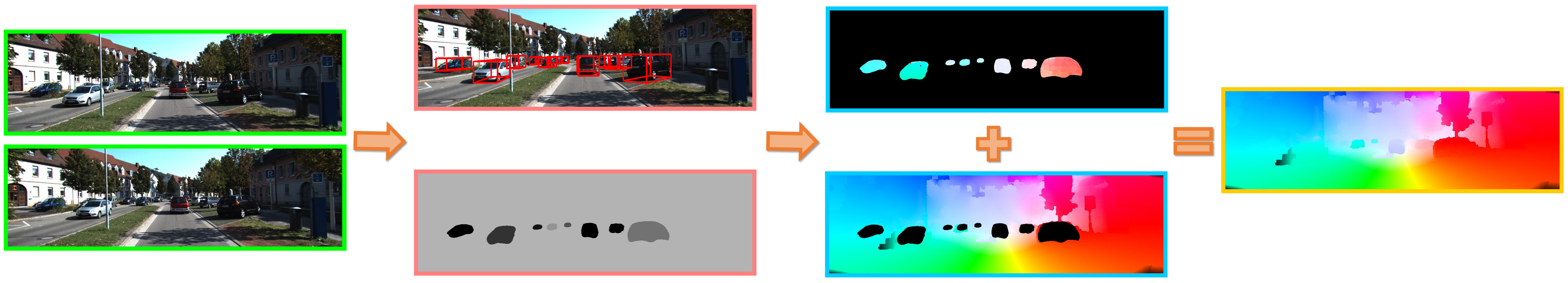}
\caption{Full pipeline of our approach. We take the input image, segment the potentially moving vehicles from the background, estimate the flow individually for every object and the background, and combine the flow for the final result.}
\label{fig:intro}
\end{center}
\end{figure*}

In contrast, we propose an alternative approach in this paper which  relies solely on exploiting  semantics to identify the potentially moving objects. 
Note that semantic segmentation is not sufficient as different vehicles might move very differently, yet form a single connected component due to occlusion. 
Instead, we exploit  instance-level segmentation, which provides us with  a different segmentation label for each vehicle. 
Given the instance segmentations, our approach then formulates the optical flow problem as a set of  epipolar flow 
estimation problems, one for each moving object. 
The background is considered as a special object whose  motion  is solely due to the ego-car. 
This  contrasts current epipolar flow approaches  \cite{Yamaguchi13,Yamaguchi14}, 
which assume a static scene wherein only the observer can move. 
As shown in our experimental evaluation, this results in much better flow estimates for moving objects. Since we formulate  the problem as a set of epipolar flow problems,  the search space is reduced from a 2D area to a 1D search along the epipolar lines. This has benefits both in terms of the computational complexity, as well as the robustness of our proposed approach. We refer the reader to Fig. \ref{fig:intro} for an illustration of our approach. 

The success of our approach relies on accurate fundamental matrix estimation for each moving object, as well as accurate matching. 
To facilitate this, our second contribution is a new convolutional net that learns to perform flow matching, and is able to estimate the uncertainty of its matches. This allows us to reject outliers, leading to better estimates for the  fundamental matrix of each moving object. 
We smooth our predictions using  semi-global block  matching \cite{Hirschmuller}, where each match from the convolutional net is restricted to lie on its epipolar line. 
We post-process our flow estimate using left-right consistency to reject outliers, followed by  EpicFlow \cite{Revaud} for the final interpolation. Additionally, we take advantage of slanted plane methods \cite{Yamaguchi14} for  background flow estimation to increase smoothness for texture-less and saturated regions. 

We demonstrate the effectiveness of our approach in the challenging KITTI 2015 flow benchmark \cite{Geiger}, and show that our approach outperforms all published approaches by a large margin. 
In the following, we first review related work, and discuss our convolutional net for flow estimation. We then present our novel approach that encodes flow as a collection of rigidly moving objects, follow by  our experimental evaluation. 


 \section{Related Work}

The classical approach for optical flow estimation involves building an  energy model, which  typically incorporates image evidence such as gradient consistency \cite{Kanade1981,Schunck1981}, warping \cite{Papenberg2006}, or matches \cite{Revaud} as unary terms. Additionally, there is a pairwise term to encourage smoothness. 
There are various methods for energy minimization and embedding of additional priors. This section summarizes several major categories.

The  study of  \cite{Black2013} shows that classical approaches to optical flow estimation are mainly gradient based methods~\cite{Kanade1981,Schunck1981}. Unfortunately, these are typically unsuitable for estimating large displacements (often encountered in traffic scenes) due to inconsistent image patch appearances. Both coarse-to-fine strategies ~\cite{Black2010} as well as inference at the original image resolution  are employed~\cite{Bruhn,Weinzaepfel}. EpicFlow \cite{Revaud} is a global approach that is very often used to interpolate sparse flow fields taking into account edges \cite{weinzaepfel:hal-00873592}. 
As shown in our experiments, its performance can be improved even further when augmented with explicit reasoning about moving objects.

Many approaches formulate flow as inference in a Markov random field   (MRF) \cite{Yang2014,Bao2014,Lempitsky,Menze,Lei}.  
 Message passing or move making algorithms are typically employed for inference. 
One of the most successful optical flow methods in the context of autonomous driving   is DiscreteFlow \cite{Menze}, which reduces the search space by utilizing only a  small number of proposals. These are shared amongst neighbors 
to increase matching performance and robustness. An MRF is then employed to encourage smoothness. 
After some post processing, the final flow is interpolated using EpicFlow \cite{Revaud}. \cite{yang2012} segments images using superpixels and approximates flow of each superpixel as homographies of 3D planes. 
Unlike our method, these methods do not exploit the fact that the background is static and only a few objects move.

Concurrent to our work,  \cite{Black_cvpr2016}  also employs semantics to help optical flow. 
In particular, they identify three classes of components: static planar background, rigid moving objects, and elements for which a compact motion model cannot be defined. A different model is then adapted for each of the three classes to refine DiscreteFlow \cite{Menze}. 
An affine transformation and a smooth deformation is fitted to moving vehicles, and homographies are fitted to planar backgrounds. In contrast, we use  a stronger 3D epipolar motion constraint for both foreground vehicles and the entire static background. 
Our experiments shows that this results in much better flow estimates. 

 In a series of papers, Yamaguchi et al. \cite{Yamaguchi13,Yamaguchi14}  exploited  epipolar constraints to reduce the correspondence  search space. 
However, they assume that the scene is static and only the camera  moves, and thus cannot handle independently moving objects. 
3D priors about the physical world have been used to estimate scene flow. 
 \cite{Vogel} assumes a piecewise planar scene and piece-wise rigid motions.  Stereo and temporal image pairs are used to track these moving planes by proposing their position and orientation. 
\cite{Geiger}  tracks independently moving objects by clustering super-pixels. However, both \cite{Geiger,Vogel} require two cameras. 

Our approach is also related to multibody flow methods (e.g.,  \cite{vidal2006,zhang2011,roussos2012}), which simultaneously segment, track, and recover structure of 3D scenes with moving objects. However, \cite{vidal2006} requires noiseless correspondences, \cite{zhang2011} uses a stereo setup, and \cite{roussos2012}  has a simple data term which, unlike our approach, does not exploit deep learning. 

Recent years have seen a rise in the application of deep learning models to low level vision. In the context of stereo,~\cite{Zbontar_2015_CVPR} uses a siamese network to classify matches between two input patches as either a match or not. Combined with smoothing, it achieves the best  performance on the KITTI stereo benchmark. Similarly,~\cite{chen2015deep} uses convolutional neural nets (CNNs) to compute the matching cost at different  scales. Different CNN architectures were investigated in \cite{zagoruyko2015learning}. Luo et al. \cite{Luo16} exploited larger context  and trained the network to produce a probability distribution over disparities, resulting in better matching. Deep learning has also been used for flow estimation  \cite{fischer2015flownet}. They  proposed a convolution-deconvolution network (i.e., FlowNet) which is  trained end-to-end, and achieves good results in real-time. 


\section{Deep Learning for Flow Estimation}

 The goal of optical flow is to estimate a 2D vector encoding the motion between two consecutive frames for each pixel. 
The typical assumption is that a local region (e.g., image patch) around each pixel will look  similar in both frames. Flow algorithms then search for the pixel displacements that produce the  best score. This process is referred to as matching.  
Traditional approaches adopt hand-crafted features, such as  SIFT \cite{SIFT}, DAISY \cite{DAISY}, census transform \cite{census} or image  gradients to represent each image patch. These are matched using a simple similarity score, e.g., via an inner product on the feature space. 
However, these features are not very robust. Flow methods based on only matching perform poorly in practice. 
To address this, sophisticated {smoothing} techniques have been developed  \cite{Yamaguchi13,Revaud,weinzaepfel:hal-00873592,Menze}. 


Deep convolutional neural networks have been shown to perform extremely well in high-level semantic tasks such as classification, semantic segmentation, and object detection.
Recently, they have been successfully trained for stereo matching  \cite{Luo16,Zbontar_2015_CVPR}, producing state-of-the-art results in the challenging KITTI benchmark \cite{kitti}. 
Following this trend, in our work we adopt a deep convolution neural network  to learn feature representations that are tailored to the optical flow estimation problem. 

\subsection{Network Architecture} 

\begin{figure*}[t]
\begin{center}
\includegraphics[width=\linewidth]{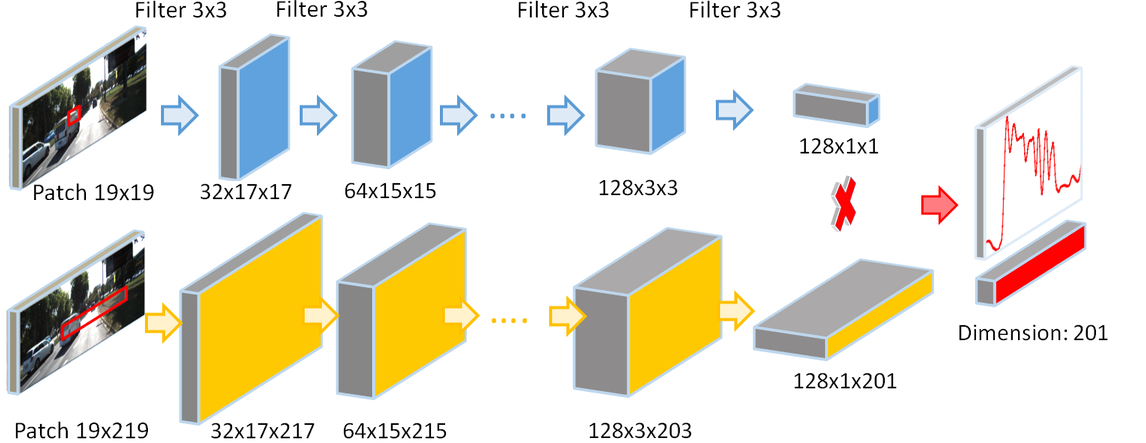}
\caption{{\bf Network Overview}: A siamese convolutional net is followed by a product layer that computes a score for each displacement. During training, for each pixel we compute a softmax over a horizontal or vertical 1D displacement, and minimize cross-entropy.
}
\label{fig:network}
\end{center}
\end{figure*}

Our network takes two consecutive frames as input, and processes them in two branches of a siamese  network to extract features. The two branches are then combined with a product layer to create a matching score for each possible displacement. We refer the reader to Fig. \ref{fig:network} for an illustration of our convolutional net. 
 
In particular, our network uses 9 convolutional layers, where each convolution is followed by  batch normalization \cite{Ioffe15} and a rectified non-linear unit (ReLU). We use $3 \times 3$   kernels  for each convolution layer. With a stride of one pixel and no pooling, this gives us a receptive field size of $19\times 19$. 
The number of filters for each convolution layer varies. As shown in Fig. \ref{fig:network}, we use the following configuration for our network: 32, 32, 64, 64, 64, 128, 128, 128. 
Note that although our network has 9 layers, the number of parameters is only around 620K. Therefore, our network is much smaller than networks used for high-level vision tasks, such as AlexNet or VGG which have 60 and 135 millions parameters, respectively.  
 As our last layer has 128 filters, the dimension of our feature vector for each pixel is also 128.

\subsection{Learning} 

To train the network, we use small image patches extracted at random from the set of pixels for which ground truth is available. This strategy is beneficial, as it provides us with a diverse set of training examples (as nearby pixels are very correlated). Furthermore, it is more memory efficient.  
Let $\cI$ and $\cI'$ be two images captured by the same camera at two consecutive times. 
Let   $(x_i,y_i)$ be the image coordinates of the center of the patch extracted at random from $\cI$, and let $(f_{x_i}, f_{y_i})$ be the corresponding ground truth flow. 
We use a patch of size $19\times 19$, since this is the size of our total receptive field. 
Since the magnitude of $(f_{x_i}, f_{y_i})$ can be very large, we create a  larger image patch in the second image $\cI'$. Including the whole search range is computationally very expensive, as this implies computing 30K scores. 
Instead, we reduce the search space and construct two training examples per randomly drawn patch, one that searches in the horizontal direction and another in the vertical direction, both centered on the ground truth point $(x+f_{x_i}, y+  f_{y_i})$. The horizontal training example is shown in Fig. \ref{fig:network}. Thus, their size is $19\times (19+R)$ and  $(19+R) \times 19$, respectively. Note that this poses no problem as we use a convolutional net. In practice, we use $R=200$. We find the network performance is not very sensitive to this hype-parameter.



 
 As we do not use any pooling  and a stride  of one,  the siamese network  outputs a single feature vector from the left branch and $(1+R)$ feature vectors from the right branch corresponding to all candidate flow locations. Note that by construction,  the ground truth  is located  in the middle of  the patch extracted in $\cI'$. 
 The matching network on top then computes the corresponding similarity score for each possible location. We simply ignore the pixels near the border of the image, and do not use them for training. 

We learn  the parameters of the model by minimizing cross entropy, where we use a soft-max over all possible flow locations.    
We thus  optimize:
$$
\min_{\mathbf{w}}\sum_{i=1}^N \sum_{s_i} p_i^{GT}(s_i)\log p_i(s_i, \mathbf{w}).
$$
where $\mathbf{w}$ are the parameters of the network, and $N$ is the total number of training examples.  $N$ is double the  number of sample patches, as we generate two training examples for each patch. In practice, we generate 22 million training examples from the 200 image pairs.
Further,   $s_i$ is the ground truth location index for patch $i$ in the second image. Recall that the second image patch was of size $(19+R)$ or $(R+19)$. Finally, $p_i^{GT}$ is the target distribution, and $p_i$ is the predicted distribution for patch $i$ according to the model (i.e., output of the soft-max). 

Note that when training neural nets, $p^{GT}$ is typically assumed to be a delta function with non-zero probability mass only for the correct hypothesis. Here, we use a more informative loss, which penalizes  depending on the distance to the ground truth configuration. We thus define
$$
p_i^{GT}(s_i) = \left\{
\begin{array}{ll}
\lambda_1&\text{if}~s_i=s_i^{GT}\\\lambda_2&\text{if}~|s_i-s_i^{GT}|=1\\\lambda_3&\text{if}~|s_i-s_i^{GT}|=2\\0&\text{o.w.}\end{array}\right..
$$
This allows the network to be less strict in discriminating patches within 3-pixels from the ground truth. In practice we choose $\lambda_1 = 0.5$, $\lambda_2 = 0.2$ and $\lambda_3 = 0.05$. 

\subsection{Inference} 
In contrast to training where we select small image patches, during inference we need to evaluate all the pixels for first image frame. 
Using the same routine  as  for learning would result in as many forward passes as the number of pixels in the image, which is computationally very expensive. 
Instead, following stereo approaches \cite{Zbontar_2015_CVPR,Luo16}, 
 we can efficiently compute the feature vector for all pixels with the siamese network using only one forward pass. Similar trick was also used when training FastRCNN \cite{girshick2015fast}, where features for all regions proposal are computed by one forward pass. 

Optical flow is more challenging than stereo matching because the search space is approximately 200 times larger, as one has to search over a 2D space. A standard searching window of size $400\times 200$ would require 300GB space to store the whole cost volume for a single image, which is prohibitive.  
Instead, we propose to use only the first top-K candidates for every location. 
This also enables the network to handle better texture-less regions as detailed in the next section. 


We perform post processing to do smoothing, handle texture-less regions as well as to better deal with occlusion and specularities. 
Toward this goal, we first utilize a simple cost aggregation to smooth the matching results, which can be noisy as the receptive field is only $19 \times 19$. 
Cost aggregation is an iterative process which, for every location $i$, updates the cost volume $c_i$ using the cost values of neighborhood locations i.e.,  
$
c_i^{t}(s_i)= \frac{\sum_{j\in \mathcal{N}(i)}c_{j}^{t-1}(s_i)}{N},\ 
$ 
where 
$\mathcal{N}(i)$ is the set of neighbor locations of $i$,   $c_i^{t}(s_i)$ is the cost volume at location $i$ during the $t$-th aggregation iteration,   $s_i$ is the flow configuration id, 
and $c_i^0(s_i)$ is the raw output from our network. Note that applying cost aggregation multiple times is equivalent to performing a weighted average over a larger neighborhood. In practice, we use 4 iterations of cost aggregation and a $5\times 5$ window size. 
Due to the fact that we only store top-K configurations  in our cost volume to reduce memory usage, neighboring locations have different sets of label ids. Thus we perform cost aggregation on the union of label sets, and  store only the top-K results after aggregation as final results $c^T_i(s_i)$. 
Note that one can interpret $c^T_i(s_i)$ as a score of the network's confidence. We thus threshold the cost $c^T_i(s_i)$ to select the most confident matches. The threshold is selected such that  on average $60\%$ of the locations are estimated as confident.
This simple thresholding on the cost aggression allows us to eliminate most specularities and shadows. In texture-less regions, the sparse top-K predicted matches sets of neighboring pixels have very little overlap. Combined with cost aggregation, scores of erroneous matches decrease through the aggregation iterations, thus eliminating erroneous matches. 
Another possible solution would be using uncertainty estimation by computing the entropy at each pixel. 
However, our experiments show that selecting top-K combined with simple thresholding works much better than  thresholding  the entropy. In practice, we used K = 30 as it balances  memory usage and performance.

\newcommand{\tp}{{\text{p}}}
\newcommand{\tP}{{\text{P}}}
\newcommand{\tw}{{\text{w}}}
\newcommand{\tu}{{\text{u}}}
\newcommand{\tv}{{\text{v}}}
\newcommand{\texto}{{\text{o}}}
\newcommand{\tq}{{\text{q}}}

\graphicspath{ {figures/} }

\section{Object-Aware Optical Flow}

In this section, we discuss our parameterization of the optical flow problem as a result of projection of 3D  scene flow. 
In particular, we assume that the world encountered in autonomous driving scenarios consists of independently moving rigid objects. The ego-car where the camera is located is a special object, which is responsible for the optical flow of the static background. 

Our approach  builds on the observation that if  the 3D motion of an object is rigid, it can be parameterized with a single transformation. 
This is captured by the fundamental matrix, which we denote by $F \in \mathcal{R}^{3 \times 3}$  with   $\text{rank} (F) = 2$. 
Let $\mathcal{I}$ and $\mathcal{I}'$ be two images captured by a single  camera 
at two consecutive times,  then for any point in a rigidly moving  object the following well-known epipolar constraint holds 
\[
	\tilde{{\tp}_i'}^{\top} F \tilde{\tp_i} = 0
\]
where ${\tp}_i = (x_i, y_i)$ and $\tp_i' = (x_i', y_i')$  are the projection of a  3D point $\tp_i$  into the two images, and $\tilde{\tp} = (x, y, 1)$ is $\tp$ in homogeneous coordinates. 
%
%
Further, the line defined by $l_i' = F_i \tilde{\tp_i}$ is the epipolar line in $\mathcal{I}'$ corresponding to point $\tp$, passing through both the epipole in $\mathcal{I}'$ and $\tp_i'$. 

\subsection{Segmenting Traffic Participants}

\begin{figure}[t]
\tiny
\begin{center}
\begin{tabular}{cc}
	\includegraphics[width=0.4\linewidth]{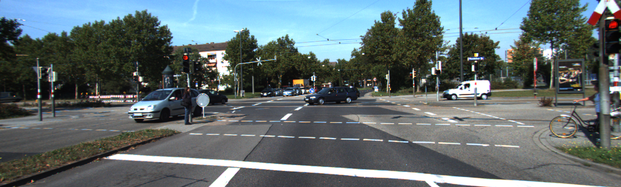} & \includegraphics[width=0.4\linewidth]{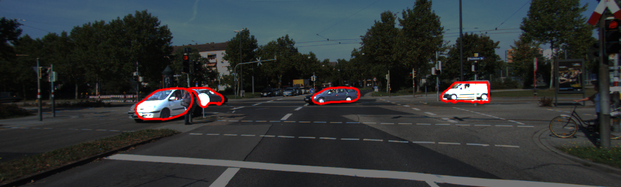} \\ 
	\includegraphics[width=0.4\linewidth]{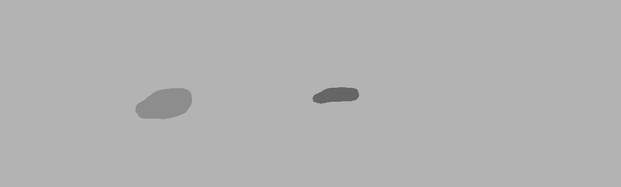} &  \includegraphics[width=0.4\linewidth]{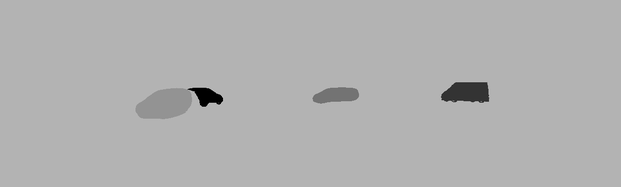} \\ 
\end{tabular}
\end{center}
\caption{\textbf{Top left}:  KITTI  image. \textbf{Top right}: Instance segmentation masks overlaid on input image. \textbf{Bottom left}: Car segmentation masks from ~\cite{ZhangCVPR2016}. \textbf{Bottom right}: Segmentation instances augmented by 3D detection \cite{XiaozhiCVPR2016} follow by CAD model fitting \cite{cadmodel}.\label{fig:masks} 
}
\end{figure}


Since only pixels belonging to one independently moving vehicle obey the same epipolar constraint, it is necessary to obtain a segmentation of the scene into independently moving objects. 
This is traditionally done by clustering the motion estimates. 
In this paper we take an alternative approach and use semantics to infer the set of potential traffic participants. Towards this goal, we exploit instance-level  segmentation which segments each traffic participant into a different component. 
Note that we aim at an upper bound on the number of moving objects, as some of the vehicles might be parked. 

To compute instance-level segmentations we exploit the approach of~\cite{ZhangCVPR2016}, which uses a multi-resolution CNN follow by a fully connected conditional random field to create a global labeling of the scene in terms of instances. 
Since only labelled training data for \emph{cars} was available, the method is unable to detect vans and trucks. This results in high precision but lower recall. 
To partially alleviate this shortcoming, we augment the instance segmentation results with  extra segmentations which are computed by performing  3D  detection \cite{XiaozhiCVPR2016} follow by CAD model fitting. In particular, we simply go over all CAD models and select the one which best aligns with the 3D box, following the technique in \cite{cadmodel}.  
Since this process has higher recall but lower precision than the instances of \cite{ZhangCVPR2016},  we only add new segmentation masks if they  do not overlap with the previously computed masks. 
We refer the reader to Fig. \ref{fig:masks} for an example. 
This process provides us with  a segmentation of the scene in terms of rigidly moving objects.  We now  discuss how to estimate flow for each moving object as well as for the background.





\subsection{Foreground Flow Estimation}

Our first goal is to reliably estimate the fundamental matrix describing the motion of each moving object. We consider this motion to be the combination of the vehicle's motion and the motion of the ego-car, that is to be the 3D motion whose projection we observe as optical flow. 
This is a challenging task, as  moving objects can be very small and  contain many specularities. 
We take advantage of the fact that our convolutional net outputs an uncertainty estimate, and only use the most confident matches for this task.  
In particular,  we use RANSAC with the 8 point algorithm \cite{Hartley97} to 
 estimate  the fundamental matrix of  each moving object independently. 
%
We then choose the hypothesis with smaller  median squared error, where   error is defined as the shortest distance between each matching point 
and its epipolar line. 

Following \cite{Yamaguchi13}, we consider the optical flow $\tu_\tp = (\tu_x, \tu_y)$ at point $\tp$ to decompose into  its rotational and translational components.  Thus
\[
	\tu_{\tp_k} = \tu_\tw(\tp_k) + \tu_\tv (\tp_k, Z_{\tp_k})
\]
where $\tu_\tw(\tp)$ is the component of the flow of pixel $\tp$ due to  rotation, and $\tu_\tv (\tp, Z_\tp)$ is the component of the flow from the translation of the object relative to the camera. Note that the direction Z here is perpendicular to the image plane of $\mathcal{I'}$. 
If the rotation is small, the rotational component can be linearized. 
We estimate the linear coefficients using matched point pairs, with the additional constraint that the point $\tp + \tu_\tw(\tp)$ must lie on the epipolar line in the second image. 

Upon application of the aforementioned linear transformation to $\mathcal{I}$, the image planes of the image patches corresponding to the object are now parallel, and related to each other only by a relative translation. This reduces the problem to either an epipolar contraction or epipolar expansion, where matching point pairs both lie on the same epipolar line. Therefore, the search for a matching point is reduced to a 1D search  along the epipolar line. The flow at a given point is then parameterized as the disparity along the epipolar line between its rectified coordinates and its matching point. 

To smooth our results, we exploit semi-global block matching (SGM)  \cite{Hirschmuller}. 
In particular, we parameterized the problem using disparity along the epipolar line as follows: 
\[
E(d) = \sum\limits_{\tp_k} C'(\tp_k, d_{\tp_k}) + \sum\limits_{\tp_k, \tp_k' \in \mathcal{N}} S(d_{\tp_k}, d_{\tp_k'})
\]
with $C'(\tp_k, d_{\tp_k})$ being the matching similarity score computed by our convolutional net with local cost aggregation to increase robustness to outliers. Note that the vz-ratio parameterization of disparity in \cite{Yamaguchi13} is unsuitable for foreground objects, as it relies on a significant relative motion in the z-direction (perpendicular to the image plane). This assumption is often violated by foreground vehicles, such as those crossing an intersection in front of the static observer. 
We use a standard smoothing term 
\[
	S(d_{\tp_k}, d_{\tp_k'}) = 
	\left\{
		\begin{array}{ll}
		\lambda_1 & \text{if } |d_{\tp_k}  - d_{\tp_k'}| = 1 \\
		\lambda_2 & \text{if } |d_{\tp_k}  - d_{\tp_k'}| > 1 \\
		0 & \text{otherwise}
		\end{array}
	\right.
\]
with $\lambda_2 > \lambda_1 > 0$. In practice, $\lambda_2 = 256$ and $\lambda_1 = 32$. 
After SGM, we use left-right consistency check to filter out outliers. The output is a semi-dense flow estimate.

Occasionally, the fundamental matrix estimation for an object fails due to either too few confident matches or too much noise in the matches.
In this case, we directly use the network's matching to obtain a flow-field. 
Finally, we use the edge-aware interpolation of EpicFlow \cite{Revaud}  to interpolate the missing pixels by performing one step of variational smoothing. This produces a fully dense flow-field for all objects. 

\subsection{Background Flow Estimation}

To estimate the background flow, we mostly follow Yamaguchi et al. \cite{Yamaguchi13}. However, we make two significant changes which greatly improve its performance. First, we restrict the matches to the areas estimated to be background by our semantic segmentation. 
We use RANSAC and the 8-point algorithm with  SIFT to estimate the fundamental matrix. 
Note that this simple approach is sufficient as background occupies most of the scene. 
Similar to the  foreground, the flow $\tu_\tp$ at a point $\tp$ is considered to be a sum of a rotational and a translational component: $\tu_\tp = \tu_\tw(\tp) + \tu_\tv (\tp, Z_\tp)$. Again, we linearize the rotational component. 
To find the matching point $\tp'$ for $\tp$, we search along the epipolar line $l'$, and parameterize the displacement vector as a scalar disparity. 


Further, the disparity at point $\tp_i$ can  be written as 
\[
d(\tp, Z_{\tp}) = |\tp + \tu_\tw(\tp) - \texto' | \frac{\frac{v_z}{Z_\tp}}{{1 - \frac{v_z}{Z_\tp}}}
\]
where $v_z$ is the forward (Z) component of the ego-motion, $\texto'$ is the epipole and $\omega_\tp = \frac{v_z}{Z_\tp}$.

We use SGM  \cite{Hirschmuller} to smooth the estimation. However, we parameterize the flow in terms of  the vz-ratio instead of directly using disparity as in the case of foreground flow estimation. 
We perform inference along 4 directions and aggregate the results. 
Finally, we post process the result by checking left-right consistency to remove outliers. 
This provides us with a semi-dense estimate of flow for the background pixels. 

Unfortunately, no matches are found by the matching pair process in occluded regions such as portions of road or buildings that disappear from view as the vehicle moves forward. An additional significant improvement over \cite{Yamaguchi13} is a 3D geometry-inspired extrapolation. Let $\delta_\tp = |\tp + \tu_\tw(\tp) - \texto' |$ be the distance between the point $\tp$ and $\texto'$. For a planar surface in the 3D world, $\delta_\tp$ is inversely proportional to $Z_\tp$. Since $v_z$ is constant for all points after the linearized rotational flow component $\tu_\tw(\tp)$ is removed, the vz-ratio is also proportional to $\delta_\tp$. 
For each point $\tp$ where the vz-ratio is not  estimated, we search along the line segment joining $\tp$ to $\texto'$ to collect a set of up to 50 vz-ratios at pixels $\tp'$, and calculate their associated $\delta_{\tp'}$. We  take advantage of semantic information to exclude points belonging to moving foreground vehicles. Using this set, we fit a  linear model which we use to estimate the  missing vz-ratio at $\tp$. 

We employ a slanted plane model similar to MotionSLIC \cite{Yamaguchi13}  to compute a dense and smooth background flow field.   This assumes that the scene is composed of small, piece-wise planar regions. 
In particular, we model the vz-ratios of the pixels in each superpixel with a plane defined as 
$\frac{v_z}{Z_\tp} = A(x-x_c) + B(y-y_c) + C$. Here, $(A,B,C)$ are the plane parameters, and  $(x_c, y_c)$ are the coordinates of the center of the superpixel. 
We simultaneously reason both about the assignments of pixels to planes, the plane parameters, and the types of boundaries between superpixels (i.e., co-planar, hinge, occlusion). Inference is simply performed by block coordinate descent. 



\begin{table}[H]
\captionsetup{font=small}
\centering
\begin{tabular}{|c|c|c|c|c|c|c|}
\hline
\multirow{2}{*}{Method} & \multicolumn{3}{c|}{Non occluded px} & \multicolumn{3}{c|}{All px}    \\ \cline{2-7} 
                        & Fl-bg      & Fl-fg      & Fl-all     & Fl-bg    & Fl-fg    & Fl-all   \\ \hline
HS   \cite{Black2013}       & 30.49 \%   & 50.59 \%   & 34.13 \%   & 39.90 \% & 53.59 \% & 42.18 \% \\ \hline
DeepFlow \cite{Weinzaepfel}    & 16.47 \%   & 31.25 \%   & 19.15 \%   & 27.96 \% & 35.28 \% & 29.18 \% \\ \hline
EpicFlow \cite{Revaud}   & 15.00 \%   & 29.39 \%   & 17.61 \%   & 25.81 \% & 33.56 \% & 27.10 \% \\ \hline
MotionSLIC  \cite{Yamaguchi13}   & 6.19 \%    & 64.82 \%   & 16.83 \%   & 14.86 \% & 66.21 \% & 23.40 \% \\ \hline
DiscreteFlow  \cite{Menze}   & 9.96 \%    & \textbf{22.17 \%}   & 12.18 \%   & 21.53 \% & \textbf{26.68 \%}  & 22.38 \% \\ \hline
SOF \cite{Black_cvpr2016} & 8.11 \% & 23.28 \% & 10.86 \% & 14.63 \% & 27.73 \% & 16.81 \% \\ \hline
Ours       & \textbf{5.75 \%}    &  22.28 \%   & \textbf{8.75 \%}    & \textbf{8.61} \% & 26.69 \% & \textbf{11.62 \%} \\ \hline

\end{tabular}
\caption{\textbf{KITTI Flow 2015 Test Set}: we compare our results with  top scoring published monocular methods that use a single image pair as input \label{test_data}} 

\end{table}

\section{Experimental Evaluation}

We evaluated our approach on the 
 KITTI Optical Flow 2015 benchmark \cite{kitti}, which consists of 200 training and 200 testing image pairs. 
 There are a number of challenges  including specularities, moving vehicles, sensor saturation,  large displacements and texture-less regions. 
 The benchmark error metric is the percentage of pixels whose error exceeds 3 px or  5 \% (whichever is greater) from the ground truth flow. We refer to (Fl-fg) and (Fl-bg) as the error evaluated only on the foreground and background pixels respectively, while  (Fl-all) denotes the average error across all pixels. 
 In this section, we first analyze our method's performance in comparison with the state-of-the-art. Additionally, we explore the impact of various stages of our  pipeline.

We trained our siamese convolutional network for 100k iterations using stochastic gradient descend with Adam \cite{kingma2014adam}. We used a batch size of 128 and an initial learning rate of 0.01 with a weight decay of $0.0005$. 
We  divided the learning rate by half at iterations 40K, 60K and  80K. 
Note that since we have 22 million training examples, the network converges before completing one full epoch. This shows that 200 images are more than enough to train the network. 
Training takes 17 hours on an NVIDIA-Titan Black GPU. However,  performance improves only slightly after 70k iterations.


\noindent{\bf Comparison to the state-of-the-art:} We first present our results\footnote{We exploit the instances of \cite{ren2016} for our submission to the evaluation server.} on the  KITTI Optical Flow 2015 test set, and compare our approach to published monocular approaches that exploit a single temporal image pair as input. 
As shown in Table \ref{test_data}, our approach significantly outperforms all published approaches. Our approach is particularly effective on the background, outperforming  MotionSLIC \cite{Yamaguchi13}. Moreover, our method's foreground performance is very close to the leading foreground estimation technique DiscreteFlow \cite{Menze}. 
As we will see in our analysis section, our method has a clear potential to exceed DiscreteFlow \cite{Menze} in foreground also. 
The test set images are fairly correlated, as many pairs are taken from the same sequence. To provide further analysis, we also computed results on the training set, where we average results over 5 folds. For each fold, 160 images are used for training, and the remaining 40 are used for testing. The same improvements as with the test set can be as seen in Table \ref{train_data}. 

\begin{table}[t]
\captionsetup{font=small}
\centering
\begin{tabular}{|c|c|c|c|c|c|c|}
\hline
\multirow{2}{*}{Method} & \multicolumn{3}{c|}{Non occluded px} & \multicolumn{3}{c|}{All px}    \\ \cline{2-7} 
                        & Fl-bg      & Fl-fg      & Fl-all     & Fl-bg    & Fl-fg    & Fl-all   \\ \hline
EpicFlow \cite{Revaud}   & 16.14 \%   & 28.75 \%   & 18.66 \%   & 27.28 \% & 31.36 \% & 28.09 \% \\ \hline
MotionSLIC  \cite{Yamaguchi13}   & 6.32 \%    & 64.88 \%   & 17.97 \%   & 15.45 \% & 65.82 \% & 24.54 \% \\ \hline
DiscreteFlow  \cite{Menze}   & 10.86 \%    & \textbf{20.24 \%}   & 12.71 \%   & 22.80 \% & \textbf{23.32 \%}  & 22.94 \% \\ \hline
Ours                 & \textbf{6.21 \%}    & 21.97 \%   & \textbf{9.35 \%}    & \textbf{9.38 \%} & 24.79 \% & \textbf{12.14 \%} \\ \hline
\end{tabular}
\caption{\textbf{KITTI Flow 2015 Training Set}: we compare our results with the state-of-the-art by averaging performance over 5 different splits of the KITTI training dataset into training/testing.   \label{train_data}} 
\end{table}

\begin{table}[h!]
\captionsetup{font=small}
\centering
\begin{tabular}{|c|c|c|c|c|c|c|}
\hline
\multirow{2}{*}{Method} & \multicolumn{3}{c|}{Non occluded px} & \multicolumn{3}{c|}{All px}    \\ \cline{2-7} 
                        & Fl-bg      & Fl-fg      & Fl-all     & Fl-bg    & Fl-fg    & Fl-all   \\ \hline
\cite{ren2016}   & \textbf{6.17} \%   & 25.30 \%   & 9.98 \%   & \textbf{9.31 \%} & 28.11 \% & 12.69 \% \\ \hline
\cite{ZhangCVPR2016}   & 6.18 \%    & 24.61 \%   & 9.82 \%   & 9.35 \% & 27.31 \% & 12.54 \% \\ \hline
\cite{ren2016} augmented with \cite{XiaozhiCVPR2016}   & \textbf{6.17} \%    & 22.06 \%   & \textbf{9.35 \%}   & \textbf{9.31 \%} & 25.08 \%  & 12.15 \% \\ \hline
\cite{ZhangCVPR2016} augmented with \cite{XiaozhiCVPR2016}       & 6.21 \%    & \textbf{21.97 \%}   & \textbf{9.35 \%}    & 9.38 \% & \textbf{24.79 \%} & \textbf{12.14 \%} \\ \hline
\end{tabular}
\caption{Flow estimation with various instance segmentation algorithms \label{mask_sources}} 
\end{table}

\begin{table}[h!]
\captionsetup{font=small}
\centering
\begin{tabular}{|l|l|l|l|l|}
\hline
             & \multicolumn{2}{l|}{Within Detected Object Masks} & \multicolumn{2}{l|}{Within Ground Truth Object Masks} \\ \hline
Method       & Non-occ px error \%      & All px error \%        & Non-occ px error \%        & All px error \%          \\ \hline
EpicFlow \cite{Revaud}    & 26.77 \%                 & 29.93 \%               & 28.75 \%                   & 31.36 \%                 \\ \hline
DiscreteFlow \cite{Menze} & 18.76 \%                 & 22.42 \%               & 20.24 \%                   & 23.32 \%                 \\ \hline
Ours         & \textbf{15.91 \%}        & \textbf{19.72 \%}      & \textbf{15.42 \%}          & \textbf{18.62 \%}        \\ \hline
\end{tabular}
\caption{Foreground flow estimation within detected object masks and within ground truth masks\label{fg_obj_mask_comparison}}
\end{table}

\begin{table}[h!]
\captionsetup{font=small}
\centering
\begin{tabular}{|c|c|c|c|}
\hline
Source for Matches for $F$ Estimation  & Non-occluded px error \% &  All px error \%  \\ \hline
EpicFlow \cite{Revaud}    &   25.02 \%           &       27.58 \%               \\ \hline
DiscreteFlow \cite{Menze}   &  23.40 \%           &       26.05 \%               \\ \hline
Our Matching Network      &   \textbf{21.97 \%}   &      \textbf{24.79 \%}     \\ \hline
\end{tabular}
\caption{Foreground flow estimation errors when F is estimated from various sources \label{F_comparison}}
\end{table}

\noindent{\bf Influence of instance segmentation:}  Table \ref{mask_sources} shows performance when using  \cite{ren2016} and  \cite{ZhangCVPR2016} to create the  instance segmentations.  
We also explore augmenting them by fitting CAD models with \cite{cadmodel} to the 3D detections of \cite{XiaozhiCVPR2016}.   
Note that a combination of segmentation and detection is beneficial. 
 A limiting factor of our foreground flow estimation performance arises when we missed moving vehicles when estimating our instances. 
Using our 5 folds on the  training set, we explore what happens when we have perfect objects masks. 
Towards this goal, we first examine the foreground flow estimation performance only on our detected vehicle masks. The left half of Table \ref{fg_obj_mask_comparison} shows that within the vehicle masks we detect, our flow estimation is significantly more accurate than our competitors in the same regions. Moreover, the right half of the same table shows that the same is true within the ground truth object masks. Thus, if the instances were further improved (e.g., by incorporating temporal information when computing them)  our method can be expected to achieve a significant improvement over the leading competitors. It is thus clear that our current bottleneck  is in the accuracy of the masks. 
\noindent{\bf Estimating Fundamental Matrix:} 
Having an accurate fundamental matrix is critical to the success of our method. While the strong epipolar constraint offers great robustness to outliers, it can also cause many problems if it is wrongly estimated. 
We now compare different matching algorithms employed to compute the  fundamental matrices, and use the rest of our pipeline to estimate flow.  
As shown in Table \ref{F_comparison}, selecting only confident matches from our network to estimate the fundamental matrix is significantly better than using the flow field estimations from other algorithms, including  DiscreteFlow. 





\newlength{\mylength}
\setlength{\mylength}{1.05cm}

\begin{figure*}[!ht]
\begin{center}
\begin{subfigure}[b]{0.250\textwidth} \centering \includegraphics[width=\textwidth]{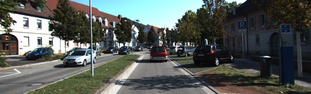} \end{subfigure}
~
\begin{subfigure}[b]{0.250\textwidth} \centering \includegraphics[width=\textwidth]{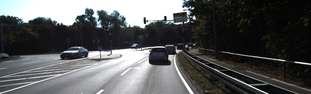} \end{subfigure}
~
\begin{subfigure}[b]{0.250\textwidth} \centering \includegraphics[width=\textwidth]{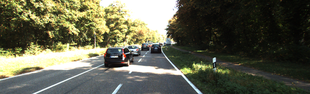} \end{subfigure} \\

\begin{subfigure}[b]{0.250\textwidth} \centering \includegraphics[width=\textwidth]{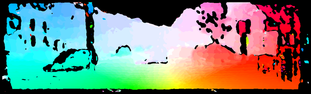} \end{subfigure}
~
\begin{subfigure}[b]{0.250\textwidth} \centering \includegraphics[width=\textwidth]{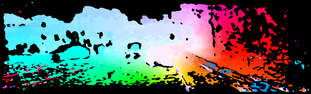} \end{subfigure}
~
\begin{subfigure}[b]{0.250\textwidth} \centering \includegraphics[width=\textwidth]{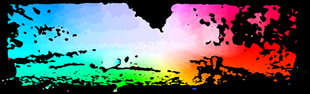} \end{subfigure} \\

\begin{subfigure}[b]{0.250\textwidth} \centering \includegraphics[width=\textwidth]{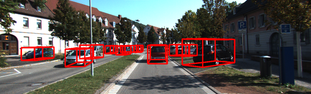} \end{subfigure}
~
\begin{subfigure}[b]{0.250\textwidth} \centering \includegraphics[width=\textwidth]{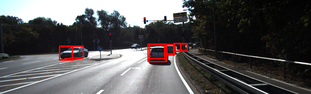} \end{subfigure}
~
\begin{subfigure}[b]{0.250\textwidth} \centering \includegraphics[width=\textwidth]{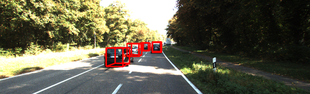} \end{subfigure} \\

\begin{subfigure}[b]{0.250\textwidth} \centering \includegraphics[width=\textwidth]{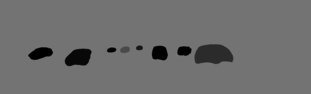} \end{subfigure}
~
\begin{subfigure}[b]{0.250\textwidth} \centering \includegraphics[width=\textwidth]{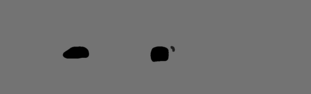} \end{subfigure}
~
\begin{subfigure}[b]{0.250\textwidth} \centering \includegraphics[width=\textwidth]{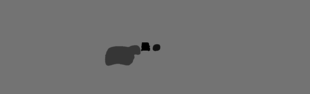} \end{subfigure} \\

\begin{subfigure}[b]{0.250\textwidth} \centering \includegraphics[width=\textwidth]{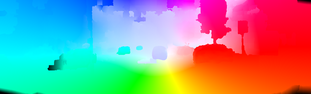} \end{subfigure}
~
\begin{subfigure}[b]{0.250\textwidth} \centering \includegraphics[width=\textwidth]{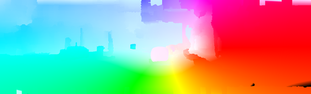} \end{subfigure}
~
\begin{subfigure}[b]{0.250\textwidth} \centering \includegraphics[width=\textwidth]{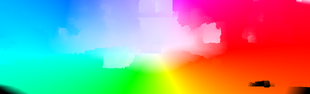} \end{subfigure} \\

\begin{subfigure}[b]{0.250\textwidth} \centering \includegraphics[width=\textwidth]{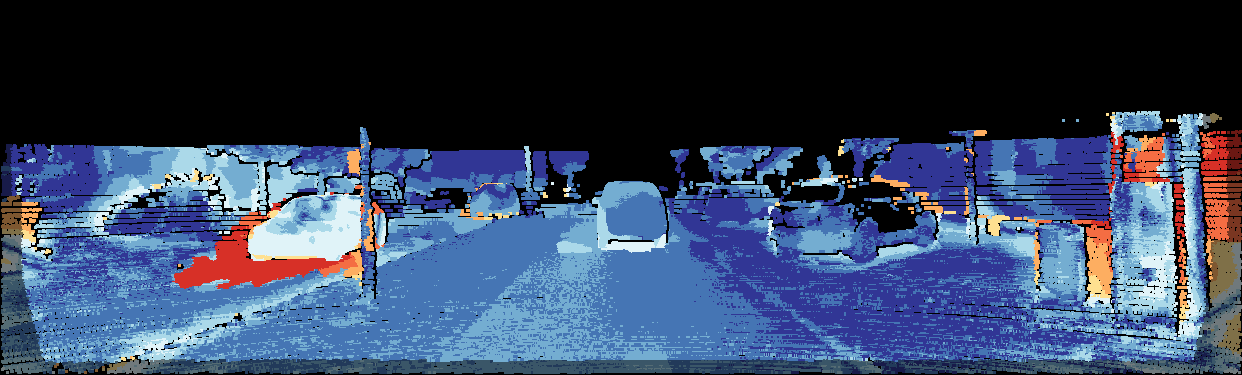}  \end{subfigure}
~
\begin{subfigure}[b]{0.250\textwidth} \centering \includegraphics[width=\textwidth]{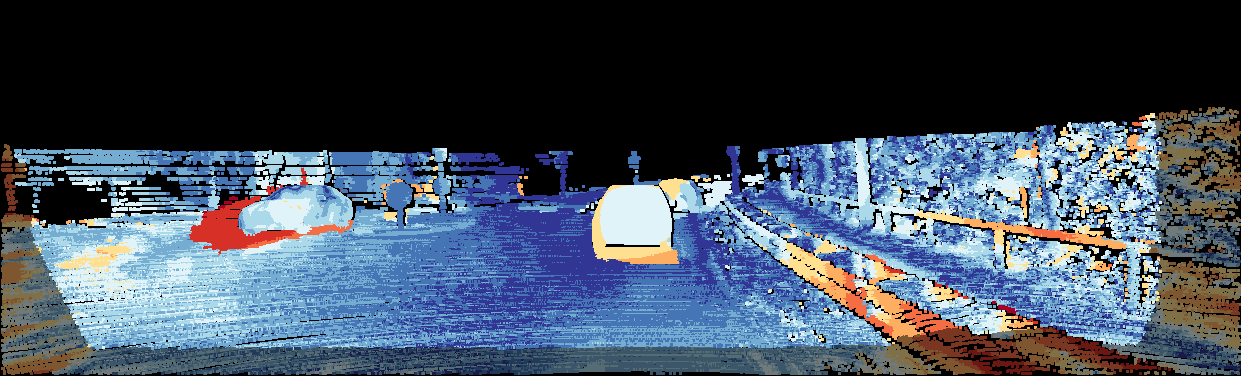} \end{subfigure}
~
\begin{subfigure}[b]{0.250\textwidth} \centering \includegraphics[width=\textwidth]{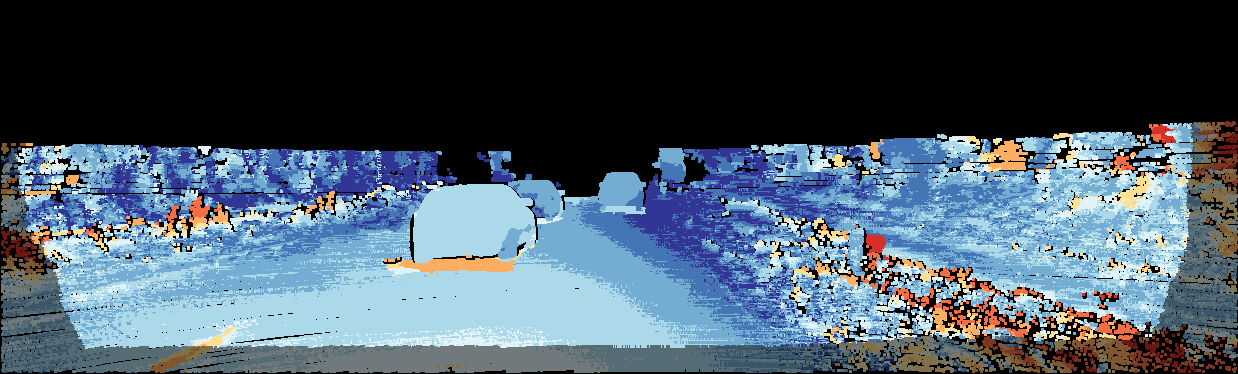} \end{subfigure} \\


\begin{subfigure}[b]{0.250\textwidth} \centering \includegraphics[width=\textwidth]{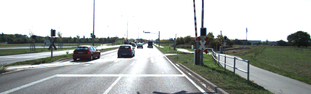} \end{subfigure}
~
\begin{subfigure}[b]{0.250\textwidth} \centering \includegraphics[width=\textwidth]{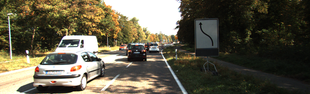} \end{subfigure}
~
\begin{subfigure}[b]{0.250\textwidth} \centering \includegraphics[width=\textwidth]{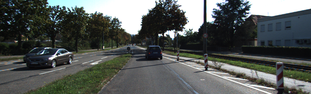} \end{subfigure} \\

\begin{subfigure}[b]{0.250\textwidth} \centering \includegraphics[width=\textwidth]{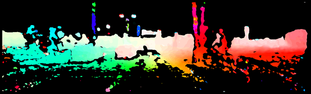} \end{subfigure}
~
\begin{subfigure}[b]{0.250\textwidth} \centering \includegraphics[width=\textwidth]{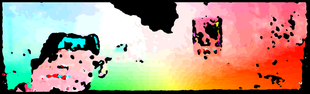} \end{subfigure}
~
\begin{subfigure}[b]{0.250\textwidth} \centering \includegraphics[width=\textwidth]{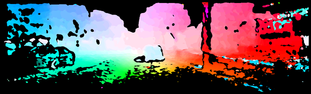} \end{subfigure} \\

\begin{subfigure}[b]{0.250\textwidth} \centering \includegraphics[width=\textwidth]{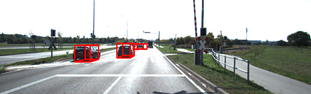} \end{subfigure}
~
\begin{subfigure}[b]{0.250\textwidth} \centering \includegraphics[width=\textwidth]{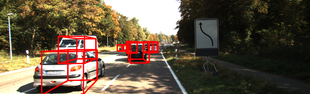} \end{subfigure}
~
\begin{subfigure}[b]{0.250\textwidth} \centering \includegraphics[width=\textwidth]{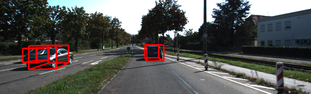} \end{subfigure} \\

\begin{subfigure}[b]{0.250\textwidth} \centering \includegraphics[width=\textwidth]{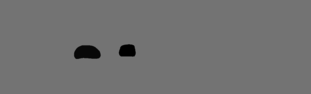} \end{subfigure}
~
\begin{subfigure}[b]{0.250\textwidth} \centering \includegraphics[width=\textwidth]{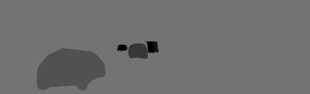} \end{subfigure}
~
\begin{subfigure}[b]{0.250\textwidth} \centering \includegraphics[width=\textwidth]{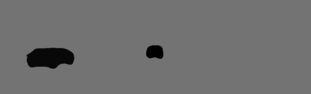} \end{subfigure} \\

\begin{subfigure}[b]{0.250\textwidth} \centering \includegraphics[width=\textwidth]{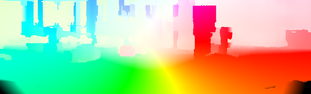} \end{subfigure}
~
\begin{subfigure}[b]{0.250\textwidth} \centering \includegraphics[width=\textwidth]{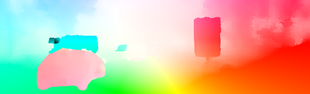} \end{subfigure}
~
\begin{subfigure}[b]{0.250\textwidth} \centering \includegraphics[width=\textwidth]{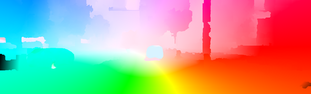} \end{subfigure} \\

\begin{subfigure}[b]{0.250\textwidth} \centering \includegraphics[width=\textwidth]{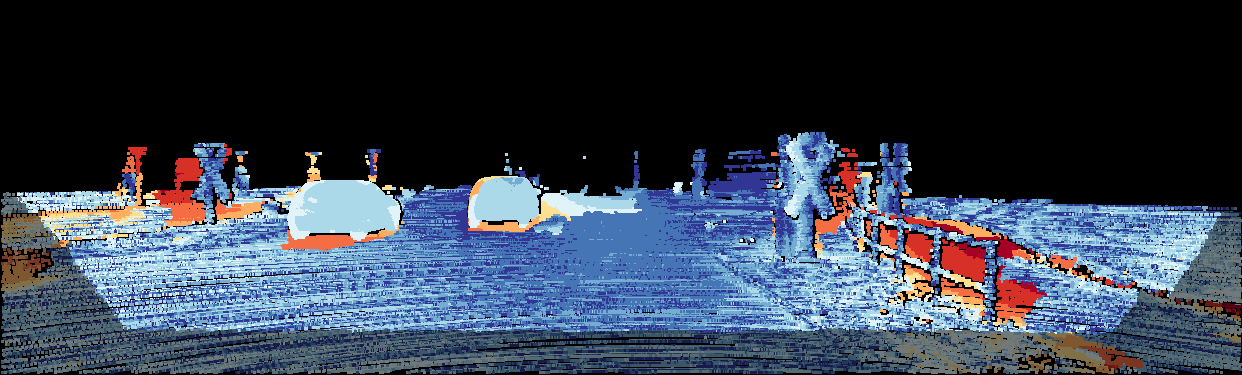} \end{subfigure}
~
\begin{subfigure}[b]{0.250\textwidth} \centering \includegraphics[width=\textwidth]{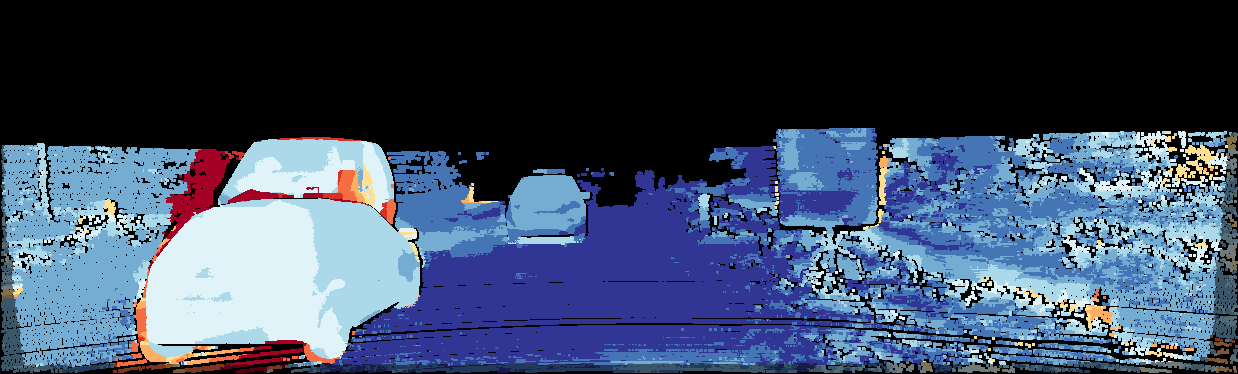} \end{subfigure}
~
\begin{subfigure}[b]{0.250\textwidth} \centering \includegraphics[width=\textwidth]{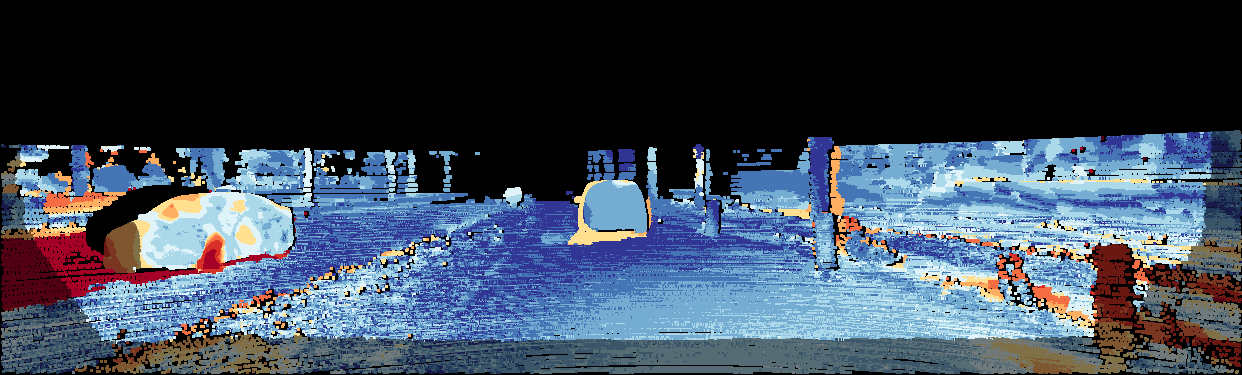} \end{subfigure}~ \\

\vspace{-0.3cm}

\caption{\small Examples of successful flow estimations. Within each group, from top to bottom: first frame of input image, confident flow produced by our network, 3D car detection results, instance segmentation output augmented by 3D car detection, final flow field, and flow field error. \label{fig:goodexample}}
\vspace{-0.5cm}

\end{center}
\end{figure*}


\noindent{\bf Qualitative Analysis:}
Fig. \ref{fig:goodexample} shows qualitative results, where each column depicts the original image, 
the network most confident estimates, the 3D detections of  \cite{XiaozhiCVPR2016}, our final instance segmentations, our final flow field and its errors. 
Our convolutional net is able to predict accurate results for most regions in the image. It leaves holes in regions including texture-less areas like the sky, occlusion due the motion of car and specularites on windshield. 
The 3D detector  is able to detect almost all cars, regardless of their orientation and size. 
Our final object masks used to label foreground objects are very accurate and contain many cars of different sizes and appearances. Note that different shades represents distinct car instances whose fundamental matrices are estimated separately. 
As shown in the last two rows, we produce very good overall performance.

\begin{figure*}[!ht]
\begin{center}

\begin{subfigure}[b]{0.250\textwidth} \centering \includegraphics[width=\textwidth]{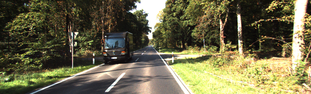} \end{subfigure}
~
\begin{subfigure}[b]{0.250\textwidth} \centering \includegraphics[width=\textwidth]{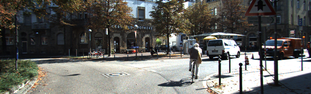} \end{subfigure}
~
\begin{subfigure}[b]{0.250\textwidth} \centering \includegraphics[width=\textwidth]{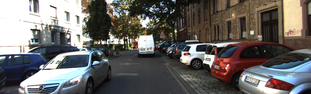} \end{subfigure} \\

\begin{subfigure}[b]{0.250\textwidth} \centering \includegraphics[width=\textwidth]{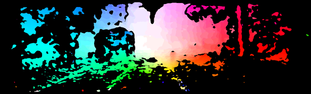} \end{subfigure}
~
\begin{subfigure}[b]{0.250\textwidth} \centering \includegraphics[width=\textwidth]{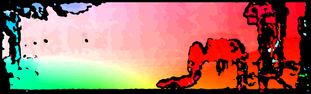} \end{subfigure}
~
\begin{subfigure}[b]{0.250\textwidth} \centering \includegraphics[width=\textwidth]{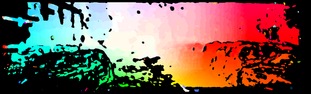} \end{subfigure} \\

\begin{subfigure}[b]{0.250\textwidth} \centering \includegraphics[width=\textwidth]{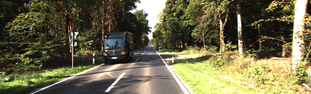} \end{subfigure}
~
\begin{subfigure}[b]{0.250\textwidth} \centering \includegraphics[width=\textwidth]{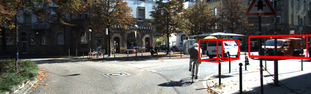} \end{subfigure}
~
\begin{subfigure}[b]{0.250\textwidth} \centering \includegraphics[width=\textwidth]{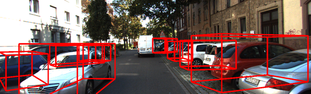} \end{subfigure} \\

\begin{subfigure}[b]{0.250\textwidth} \centering \includegraphics[width=\textwidth]{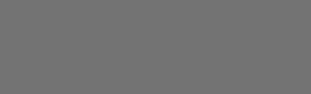} \end{subfigure}
~
\begin{subfigure}[b]{0.250\textwidth} \centering \includegraphics[width=\textwidth]{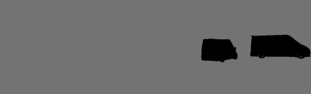} \end{subfigure}
~
\begin{subfigure}[b]{0.250\textwidth} \centering \includegraphics[width=\textwidth]{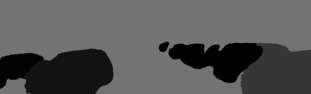} \end{subfigure} \\

\begin{subfigure}[b]{0.250\textwidth} \centering \includegraphics[width=\textwidth]{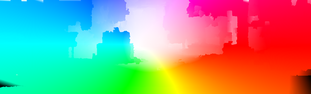} \end{subfigure}
~
\begin{subfigure}[b]{0.250\textwidth} \centering \includegraphics[width=\textwidth]{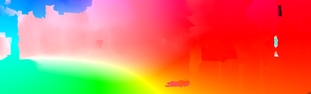} \end{subfigure}
~
\begin{subfigure}[b]{0.250\textwidth} \centering \includegraphics[width=\textwidth]{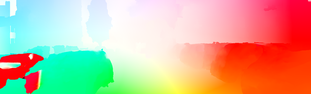} \end{subfigure} \\

\begin{subfigure}[b]{0.250\textwidth} \centering \includegraphics[width=\textwidth]{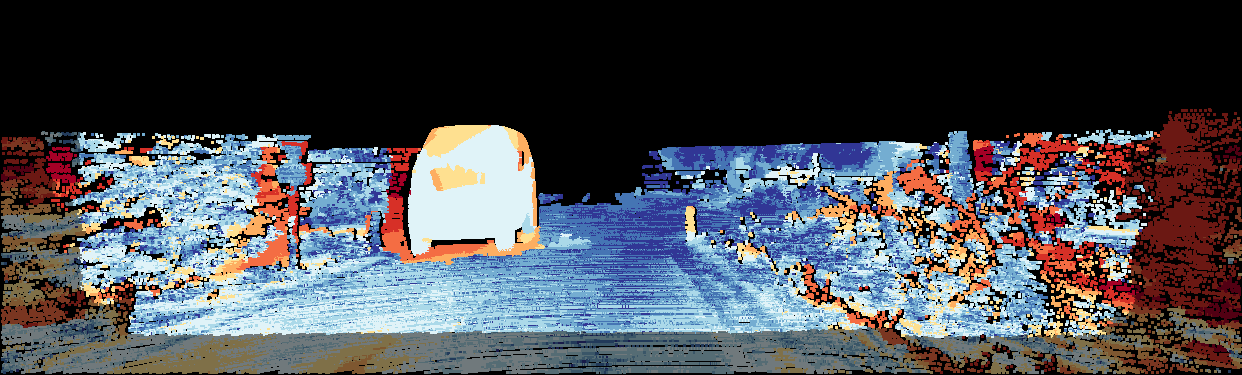} \caption{\tiny} \end{subfigure}
~
\begin{subfigure}[b]{0.250\textwidth} \centering \includegraphics[width=\textwidth]{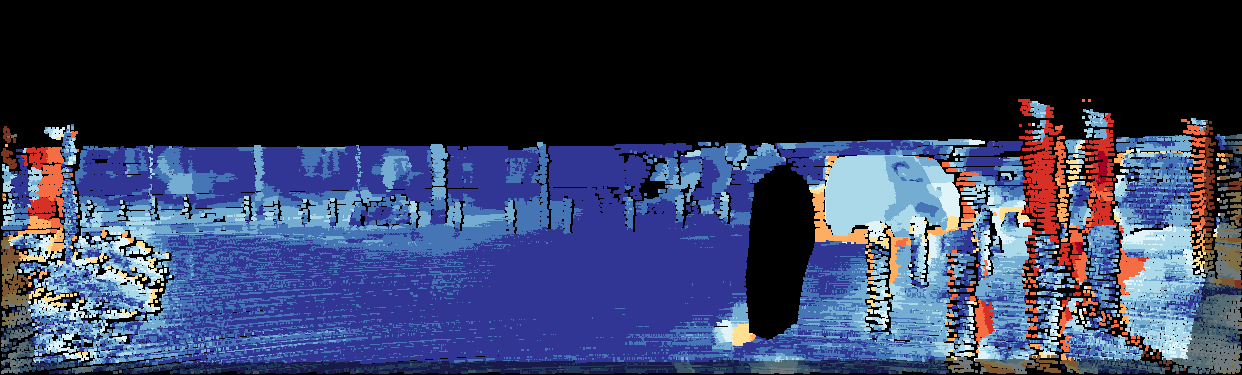} \caption{\tiny} \end{subfigure}
~
\begin{subfigure}[b]{0.250\textwidth} \centering \includegraphics[width=\textwidth]{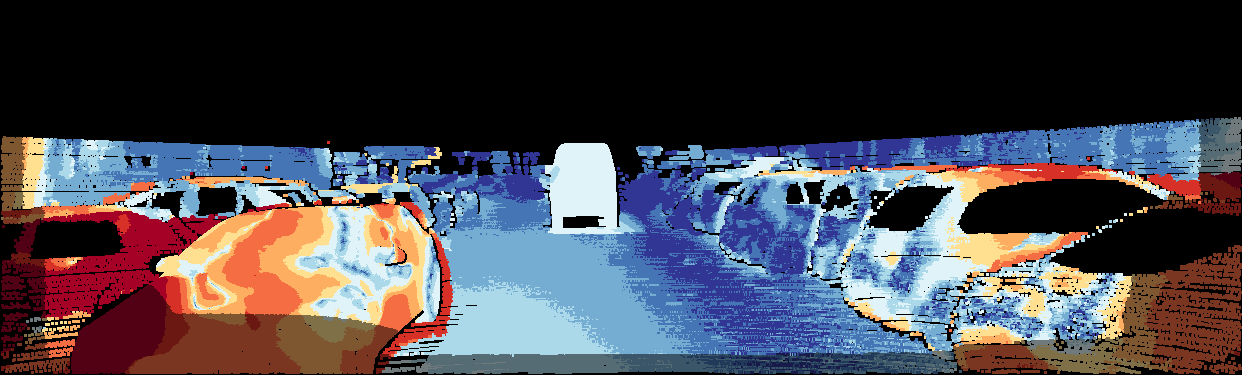} \caption{\tiny} \end{subfigure} \\

\vspace{-0.250cm}
\caption{\small Failure cases for our algorithm. }
\label{fig:bad_cases}
\vspace{-0.5cm}
\end{center}
\end{figure*}

\noindent{\bf Failure Modes:} Our technique has several failure modes. 
If a car is not segmented, the estimation of flow defaults to using the epipolar constraint of the background. This happens particularly often with trucks and vans, as we do not have training examples of these types of vehicles to train our segmentation and detection networks. Fig. \ref{fig:bad_cases}(a) shows an example where a van is not segmented. By coincidence, its true epipolar lines are almost identical with those calculated using the background fundamental matrix. As such, its flow estimation is still mostly correct. 
If object masks contain too many background pixels (which are outliers from the perspective of foreground fundamental matrix estimation), our algorithm can also fail. This is commonly associated with objects identified by the 3D object detector rather than the instance-segmentation algorithm, as the 3D detection box might be misaligned with the actual vehicle. Moreover, fitting CAD models to monocular images is not a trivial task. The right-most car in Fig. \ref{fig:bad_cases}(b) is such an example. 
The other failure mode of our approach is wrong estimation of the fundamental matrix, which can happen when the matches are very sparse or contain many outliers. Fig \ref{fig:bad_cases}(c) shows such an example, where the fundamental matrix of the left-most car is incorrectly estimated due to the sparseness in confident matching results (showing in the second row).



\section{Conclusion}

We have proposed an approach to monocular flow estimation in the context of autonomous driving which builds on the observation that the scene is composed of a static background as well a relatively small number of traffic participants (including the ego-car). 
We have shown how instance-level segmentation and 3D object detection can be used to segment the different vehicles, and proposed a new convolutional network that can accurately match patches. 
Our experiments showed that we can outperform the state-of-the-art by a large margin in the challenging KITTI 2015 flow benchmark. In the future, we plan to estimate the  different traffic participants by reasoning temporally when doing instance-level segmentation.  
\\

\noindent{\bf Acknowledgements: } This work was partially supported by ONR-N00014-14-1-0232, Samsung, and NSERC. 

\bibliographystyle{splncs}
\bibliography{egbib}
\end{document}